# Sentiment Analysis of Covid-related Reddits

Yilin Yang  
University of Ottawa  

Tomas Fieg  
University of Ottawa  

Marina Sokolova  
IBDA@ Dalhousie University,  
and University of Ottawa  

yyang202@uottawa.ca   tfieg050@uottawa.ca   sokolova@uottawa.ca

**Abstract.** This paper focuses on Sentiment Analysis of Covid-19 related messages from the r/Canada and r/Unitedkingdom subreddits of Reddit. We apply manual annotation and three Machine Learning algorithms to analyze sentiments conveyed in those messages. We use VADER and TextBlob to label messages for Machine Learning experiments. Our results show that removal of shortest and longest messages improves VADER and TextBlob agreement on positive sentiments and F-score of sentiment classification by all the three algorithms.

## 1   Introduction

The global Covid-19 pandemic has changed day-to-day life of the world population. Online platforms such as Reddit, Instagram, Twitter, Facebook, Snapchat, Youtube, provided a space for people to discuss those changes, voice opinions, and share perspectives of the pandemic. With 10,000-character limit for a comment, Reddit provides an ample material for in-depth Sentiment Analysis. Our current study focuses on Covid-19 comments posted on r/Canada and r/Unitedkingdom.

We apply VADER and TextBlob to label texts with either positive or negative sentiments. Our manual annotation shows a strong support for the negative sentiment labels. We use Random Forest, Naive Bayes, and Support Vector Machine to classify messages assigned with the same sentiments by both tools.

We investigate an impact of corpus variability within a single set on sentiment labelling and sentiment classification. Specifically, we focus on messages' length variability. To the best of our knowledge, this is a novel study that has not been carried out before. Our results show that removal of rarely occurred events from the data, i.e., the shortest and longest messages, 10 % of all the messages, increases agreement between VADER and TextBlob on positive sentiments by 5% (UK) and 6% (Canada) and improves F-scores obtained by all three algorithms on both the Canada set and the UK sets.

## 2 Related Work

Sentiment Analysis of the online posts helps to understand how Covid-19 has affected personal lives and sense of well-being of the worldwide population. Below we list papers most closely related to our work. For additional references we refer to (Tsao et al, 2021).

In recent years, Reddit has attracted a growing number of Sentiment Analysis studies. Sawicki et al (2021) show a high usability of the data for topic analysis, information processing, linguistic analysis, etc. Masood et al (2020) compared manual annotation of the Reddit anorexia-related messages and their sentiment classification. The authors concentrated on labels' validity. Our study, instead, focuses on the impact of the message length data uniformity on sentiment classification results.

Sentiment analysis of the Covid-related posts from r/Depression (Chen & Sokolova, 2021) focused on messages united by the same content of depression. We do not impose a united content requirement and instead focus on subreddits from two countries, i.e., Canada and UK.

Jalil et al. (2022) analyzed emotions expressed toward Covid-19 by applying Multi-depth DistilBERT to Twitter messages. Although showing a high performance on Twitter data, transformer-based algorithms do not perform well on longer messages, as their capacity diminishes after 512 characters. To accommodate Reddit messages, that tend to have longer texts, we use three supervised machine learning algorithms, Naive Bayes, Support Vector Machine, and Random Forest.

Many studies connect lexicographical properties of online messages with sentiment analysis (Larsson and Ljungberg, 2021; Sokolova & Bobicev, 2018). At the same time, few studies focus on variability within a single set of online messages or address low-frequent data (Gries, 2006). We bridge this gap by studying impact of messages' length variability on sentiment analysis of the data. We show that removal of shortest and longest messages increases agreement between VADER and TextBlob and improves sentiments classification results.

## 3 Reddit Datasets

Reddit[1] attracts a large following of younger, socially mobile, and active people whose lifestyle was significantly affected by the Covid pandemic and ensued restrictions (Chen and Sokolova, 2021). In this study, we worked with r/Canada and r/Unitedkingdom subreddits. We collected comments posted from January 1, 2021, to June 30, 2021. Thereafter, we filtered for Covid-19 messages using these keywords: lockdown, pandemic, coronavirus, quarantine, covid, vaccine, first dose, second dose, third dose, booster, vaccination, first shot, second shot, and third shot. After that we removed bot messages.
Figure 1 presents word-wise distribution of messages in two sets.

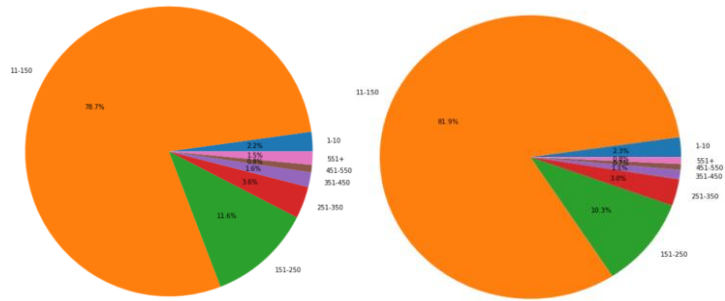

*Figure 1: Canada and UK data sets, length in words*

We identified messages-outliers as messages with 10 words or less and comments with 250 and more words. They jointly constitute less than 10% of the data. The remaining messages, 11-249 words long, constitute > 90% of the data sets. Table 1 reports on the data sets.

*Table 2: r/Canada and r/Unitedkingdom data sets.*

| Jan1 – June 30, 2021 | **Canada** | **UK** |
|---|---|---|
| Contributors | 1226363 | 785530 |
| Messages posted | 867418 | 732094 |
| All COVID -related messages | 9232 | 4218 |
| Messages, < 11 words | 201 | 95 |
| Messages, > 250 words | 702 | 237 |
| Covid-related messages, 11- 249 words | 8336 | 3886 |

---

[1] https://www.reddit.com

We label the data sets with two sentiment labeling tools, VADER[2] and TextBlob[3]. Vader is a lexicon and rule-based sentiment analysis tool attuned to sentiments expressed in social media. TextBlob is a Python library for processing textual data. It provides an API for sentiment labeling with pre-trained models.

Vader and TextBlob have a relatively low agreement among the assigned sentiment labels. For all the Covid-related messages, only 62.17% of the Canadian messages and only 60.95% of the UK messages have the same sentiment labels. Removal of messages-outliers increases the tools' agreement on positive labels: from 35.35% to 41.32% of agreed upon positive labels for the Canada set and from 29.95% to 35.77% for the UK set. However, agreement on negative messages remains stable. We report the labelling details in Table 2.

Table 2: VADER and TextBlob sentiment labelling of the data sets

| | All Covid-related messages | | | |
|---|---|---|---|---|
| | **Canada** | | **UK** | |
| | Positive | Negative | Positive | Negative |
| Consistent/Agreed | 3947 | 1799 | 1570 | 1001 |
| Inconsistent | 3493 | | 1647 | |
| Total | 9239 | | 4218 | |
| **Covid-related messages, 11-249 words** | | | | |
| | **Canada** | | **UK** | |
| | Positive | Negative | Positive | Negative |
| Consistent/Agreed | 3444 | 1704 | 1390 | 956 |
| Inconsistent/Disagreed | 3188 | | 1540 | |
| Total | 8336 | | 3886 | |

## 4 Manual Analysis

We use unsupervised manual annotation of a complete message text. In this scheme, annotators use their own judgement to assign either positive or negative sentiment to a message. We pick up messages at random, to minimise a possible bias of selection.
Two annotators receive the same group of messages and assign one of two mutually exclusive labels to a message. They work independently,

---

[2] https://vadersentiment.readthedocs.io/en/latest/
[3] https://pypi.org/project/textblob/

and do not influence each other's classification. The annotators discuss the results afterwards. We calculate Agreement between the annotators (Eq. 1).

$$Agreement = \frac{\text{\# of labels on which annoators agree}}{\text{all messages in the group}}$$

*Eq. 1.: Inter-annotation' agreement (%).*

We start with annotation of messages before they were labelled by VADER and TextBlob. We randomly select 60 messages, 30 messages for Canada and 30 messages for UK. The results show significant agreement for negative sentiments (Table 3, top). Next, we label messages with VADER and TextBlob and manually assess validity of negative labels. We randomly select two groups of 50 messages from the Canada and UK messages that VADER and TextBlob both labelled as negative (Table 3, middle). On the next step, we recalculate agreements for the two groups, keeping messages 11-249 words long. Manual annotations support negative labels assigned simultaneously by VADER and TextBlob. From 100 messages, only 2 messages are perceived positive by both annotators. At the same time, both annotators agree that 89 messages are indeed negative (Table 3, bottom).

*Table 3: Inter-annotator agreement on comments.*

| **60 unlabelled comments** | | | | |
|---|---|---|---|---|
| | Canada – 30 messages | | UK- 30 messages | |
| Both agree: | positive | negative | positive | negative |
| | 5 | 14 | 6 | 18 |
| Agreement | 0.633 | | 0.800 | |
| **100 comments labelled negative by VADER and TextBlob** | | | | |
| | Canada – 50 comments | | UK – 50 comments | |
| Both agree: | positive | negative | positive | negative |
| | 2 | 43 | 0 | 46 |
| Agreement | 0.90 | | 0.92 | |
| **95 comments, 11- 249 words, labelled negative by VADER and TextBlob** | | | | |
| | Canada – 49 comments | | UK – 46 comments | |
| Both agree: | positive | negative | positive | negative |
| | 2 | 42 | 0 | 43 |
| Agreement | 0.90 | | 0.94 | |

The annotators noticed commonalities and differences among topics of negative sentiments found in the two data sets. In r/Canada messages, negative sentiments convey vaccine hesitancy, doubts in government policies, esp. failure with limiting deaths, doubts in Covid-19 statistics and reported risk of vaccines compared to risk of covid, and concerns of covid spread due to travel. In r/Unitedkingdom messages, negative sentiments reveal vaccine hesitancy, doubts in government-imposed restrictions and how they affect the work-life balance, covid cases in hospitals and their effect on staff, and government failure with lockdown and Brexit.

The manual analysis has revealed that the Reddit users express their feelings in various ways, including sarcasm. They often use sarcasm in longer comments, where sarcastic paragraphs can be included as parts of messages. Those sarcastic expressions can impede sentiment classification, as users may write their comment using words that the machine learning algorithms interpret as positive while they convey their negative feelings in the comment.

## 5 Sentiment Classification

Binary Sentiment Classification classifies text as either positive or negative (Parikh, 2020). The goal is to create a machine learning algorithm that can predict the sentiment of a previously unseen text, as either positive or negative. To accomplish this, we use three machine learning algorithms: multinomial Random Forest (RF), Naive Bayes (NB), and Support Vector Machines (SVM).[4] We tokenize the data and build two representations: Bag of Words (BOW) and Term Frequency-Inverse Document Frequency (TF-IDF).

We classify positive and negative sentiments in four sets: Canada, all Covid related messages, or 3947 positive and 1799 negative messages; UK, all Covid-related messages, or 1570 positive and 1001 negative messages; Canada, Covid-related messages, 11-249 words long, or 3444 positive and 1704 negative messages; and UK, Covied-related messages, 11-249 words long, or 1390 positive and 956 negative messages. The messages are assigned with the same sentiment by VADER and TextBlob.

---

[4] http://www.weka.io

We use 10-fold cross-validation to find a classifier that achieves the highest Macro F-scores. The three algorithms achieve a higher F-score when using BOW. Our results improve after we reduce the text noise by removing low-frequency words from the data representation. We report the best obtained results for each algorithm in Tables 4 and 5.

*Table 4: The best sentiment classification results of all the Covid-related messages; BOW representation.*

| | Canada (3947 positive and 1799 negative messages) | | | | |
|---|---|---|---|---|---|
| | Min. word frequency | Precision | Recall | F-score | Accuracy |
| RF | 9 | 0.835 | 0.679 | 0.749 | 0.791 |
| NB | 3 | 0.773 | 0.788 | 0.780 | 0.804 |
| SVM | 1 | 0.855 | 0.832 | 0.844 | 0.868 |
| | UK (1570 positive and 1001 negative messages) | | | | |
| | Min. word frequency | Precision | Recall | F-score | Accuracy |
| RF | 8 | 0.807 | 0.748 | 0.776 | 0.789 |
| NB | 6 | 0.796 | 0.795 | 0.795 | 0.806 |
| SVM | 8 | 0.810 | 0.789 | 0.799 | 0.812 |

*Table 5: The best sentiment classification results of the Covid-related messages, 11-249 words; BOW representation.*

| | Canada (3444 positive and 1704 negative messages) | | | | |
|---|---|---|---|---|---|
| | Min. word frequency | Precision | Recall | F-score | Accuracy |
| RF | 8 | 0.803 | 0.726 | 0.763 | 0.799 |
| NB | 4 | 0.801 | 0.806 | 0.803 | 0.824 |
| SVM | 6 | 0.855 | 0.835 | 0.845 | 0.866 |
| | UK (1390 positive and 956 negative messages) | | | | |
| | Min. word frequency | Precision | Recall | F-score | Accuracy |
| RF | 6 | 0.799 | 0.795 | 0.788 | 0.795 |
| NB | 5 | 0.801 | 0.803 | 0.802 | 0.808 |
| SVM | 8 | 0.817 | 0.804 | 0.810 | 0.818 |

Note that SVM has shown the best performance on both sets. For both sets of experiments, SVM uses a linear polynomial with softmargin 0.3 for the Canada set and a linear polynomial with softmargin 0.4 for the UK set.

The obtained results show that reduction in data variability improves Machine Learning results. We achieve that through expulsion of low-frequent events such as shortest and longest messages. This improvement is more pronounced in the Recall and F-score results. Absent shortest and longest messages, or 10 % of all the messages, the three algorithms increase their ability to recognize true positive and true negative sentiments. Their confidence in those sentiments, expressed by Precision, also increases albeit on a smaller scale.

**7 Conclusions and Future Work**

In this work we have shown that variability of online message sets impacts results of binary sentiment classification of those messages. Specifically, we concentrated on reduction in messages' length variability by removal of low-frequent events such as shortest and longest messages rarely appeared in the sets.

Our empirical results were obtained on Reddit messages gathered from r/Canada and r/Unitedkingdom subreddits. We collected messages posted from January 1, 2021, to June 30, 2021. We extracted Covid-related comments and then classified them into positive or negative sentiment categories. Our manual annotation has shown a high agreement on negative sentiments for both r/Canada and r/Unitedkingdom comments. Common topics in negative sentiments are vaccine hesitancy and doubts in government policies.

We have shown that VADER and TextBlob, sentiment labelling tools, improved their positive sentiment agreement by 5%-6% after the shortest and the longest messages, approx. 10% of the data *in toto*, were removed from the data. This is valid for both Canada and UK messages.

Our Machine Learning results also have improved after reduction in data variability. All three algorithms, RF, NB, and SVM, better recognized true positive and true negative sentiments that resulted in higher

Recall and F-score results. This result is valid for both Canada and UK messages.

For future work, we envision a more detailed study on relationship between variability of data sets and the sets' sentiment classification. Another direction is to study sarcasm in Covid-related messages and its impact on sentiment classification.

**Bibiliography**